%% file: main.tex
\documentclass[lettersize,journal]{IEEEtran}
\usepackage{amsmath,amsfonts}
\usepackage{algorithmic}
\usepackage{array}
\usepackage{textcomp}
\usepackage{stfloats}
\usepackage{url}
\usepackage{verbatim}
\usepackage{graphicx}
\def\BibTeX{{\rm B\kern-.05em{\sc i\kern-.025em b}\kern-.08em
    T\kern-.1667em\lower.7ex\hbox{E}\kern-.125emX}}
\usepackage{balance}
\usepackage[font=footnotesize]{caption}
\usepackage{anyfontsize}
\usepackage{cite}
\usepackage{graphicx}
\usepackage{subcaption}
\usepackage{hyperref}
\usepackage{amsmath,amssymb,amsfonts}
\usepackage{xcolor}
\def\BibTeX{{\rm B\kern-.05em{\sc i\kern-.025em b}\kern-.08em
    T\kern-.1667em\lower.7ex\hbox{E}\kern-.125emX}}
\usepackage{float}
\usepackage[normalem]{ulem}
\usepackage{multirow}
\usepackage{soul}
\usepackage{tabularray}
\UseTblrLibrary{diagbox}
\usepackage{graphics}
\usepackage{algorithm2e}
\usepackage{adjustbox}
\usepackage{hyperref}

\usepackage{cuted} 
\usepackage{booktabs}
\usepackage{ragged2e}
\usepackage{colortbl}
\providecommand{\DIFadd}[1]{{\protect\color{black}{#1}}} %DIF PREAMBLE
\providecommand{\DIFdel}[1]{{\protect\color{red}\sout{#1}}}                      %DIF PREAMBLE

\begin{document}
\title{SAFE-RL: Saliency-Aware  Counterfactual Explainer for Deep Reinforcement Learning Policies}
\author{Amir Samadi, Konstantinos Koufos, Kurt Debattista and Mehrdad Dianati
% \vspace{-0.4cm}
\thanks{This research is sponsored by Centre for Doctoral Training to Advance the Deployment of Future Mobility Technologies (CDT FMT) at the University of Warwick. The authors are affiliated with WMG at the University of Warwick, Coventry CV4 7AL, UK (email: amir.samadi, konstantinos.koufos, k.debattista, and m.dianati@warwick.ac.uk). Mehrdad Dianati is also affiliated with the EEECS at Queen’s University Belfast, Belfast BT7 1NN, UK.}}

\markboth{Journal of \LaTeX\ Class Files,~Vol.~18, No.~9, September~2020}%
{How to Use the IEEEtran \LaTeX \ Templates}

\maketitle
\thispagestyle{empty}
\pagestyle{empty}

\begin{abstract}
While Deep Reinforcement Learning (DRL) has emerged as a promising solution for intricate control tasks, the lack of explainability of the learned policies impedes its uptake in safety-critical applications, such as automated driving systems (ADS). Counterfactual (CF) explanations have recently gained prominence for their ability to interpret black-box Deep Learning (DL) models. CF examples are associated with minimal changes in the input, resulting in a complementary output by the DL model. Finding such alternations, particularly for high-dimensional visual inputs, poses significant challenges. Besides, the temporal dependency introduced by the reliance of the DRL agent action on a history of past state observations further complicates the generation of CF examples. To address these challenges, we propose using a saliency map to identify the most influential input pixels across the sequence of past observed states by the agent. Then, we feed this map to a deep generative model, enabling the generation of plausible CFs with constrained modifications centred on the salient regions. We evaluate the effectiveness of our framework in diverse domains, including ADS, Atari {\color{black}{Pong, Pacman and space-invaders}} games, using traditional performance metrics such as validity, proximity and sparsity.  Experimental results demonstrate that this framework generates more informative and plausible CFs than the state-of-the-art for a wide range of environments and DRL agents. In order to foster research in this area, we have made our datasets and codes publicly available~\url{https://github.com/Amir-Samadi/SAFE-RL}.
\end{abstract}


\section{Introduction}
\label{sec:intro}
% \input{tex/intro}
\input{tex/intro_related}

 \section{Saliency-Aware CF DRL Explainer}
 \label{section:method}

\input{tex/method}
 \section{Evaluations}
 \label{section:results}

\input{tex/results}

 \section{Conclusions}
 \label{section: conlusion}

\input{tex/conclusion}

\bibliography{ref.bib}

\end{document}

%% file: tex/intro_related.tex
% The necessity of explainability for DRL
%\st{Deep Reinforcement Learning (DRL) has emerged as a powerful decision-making technique in environments with high-dimensional input states where traditional RL algorithms do not perform well}. 
Deep Reinforcement Learning (DRL) has emerged as a powerful decision-making technique for solving intricate control tasks across diverse domains, ranging from robotics~\cite{zhu2021deep} to gaming~\cite{berner2019dota}. This achievement is facilitated by employing Deep Neural Networks (DNNs) that can effectively learn to map high-dimensional state spaces to actions forming the policy of the DRL agent. However, the lack of interpretability in the operation of DNNs and subsequently in the learned policies hinders the application of DRL agents in safety-critical scenarios, such as automated/autonomous driving systems (ADS), where understanding the decision-making process is indispensable~\cite{olson2021counterfactual}. The explicability of the DRL policies becomes paramount, as it fosters trust and empowers human operators to intervene, debug, and ensure transparency of the system's behaviour. As a result, such explanations can aid diverse stakeholders, including end-users, engineers, and legal authorities in understanding the decision-making process that governs the operation of DRL-based systems~\cite{wachter2017counterfactual}. To this end, we investigate counterfactual (CF) explanations, a promising method for eXplainable Artificial Intelligence (XAI) that has recently gained momentum in interpreting the decision-making of black-box DNNs~\cite{samadi2023counterfactual}. A CF explainer entails applying minimal modifications to the input that result in an alternative output from the deep learning model. Comparing the understandable and actionable CF examples with the original state-action pairs can bridge the gap between opaque DRL policies and human-interpretable reasoning.
\begin{figure}
\centering
\includegraphics[width=0.91\linewidth, height=4.1cm]{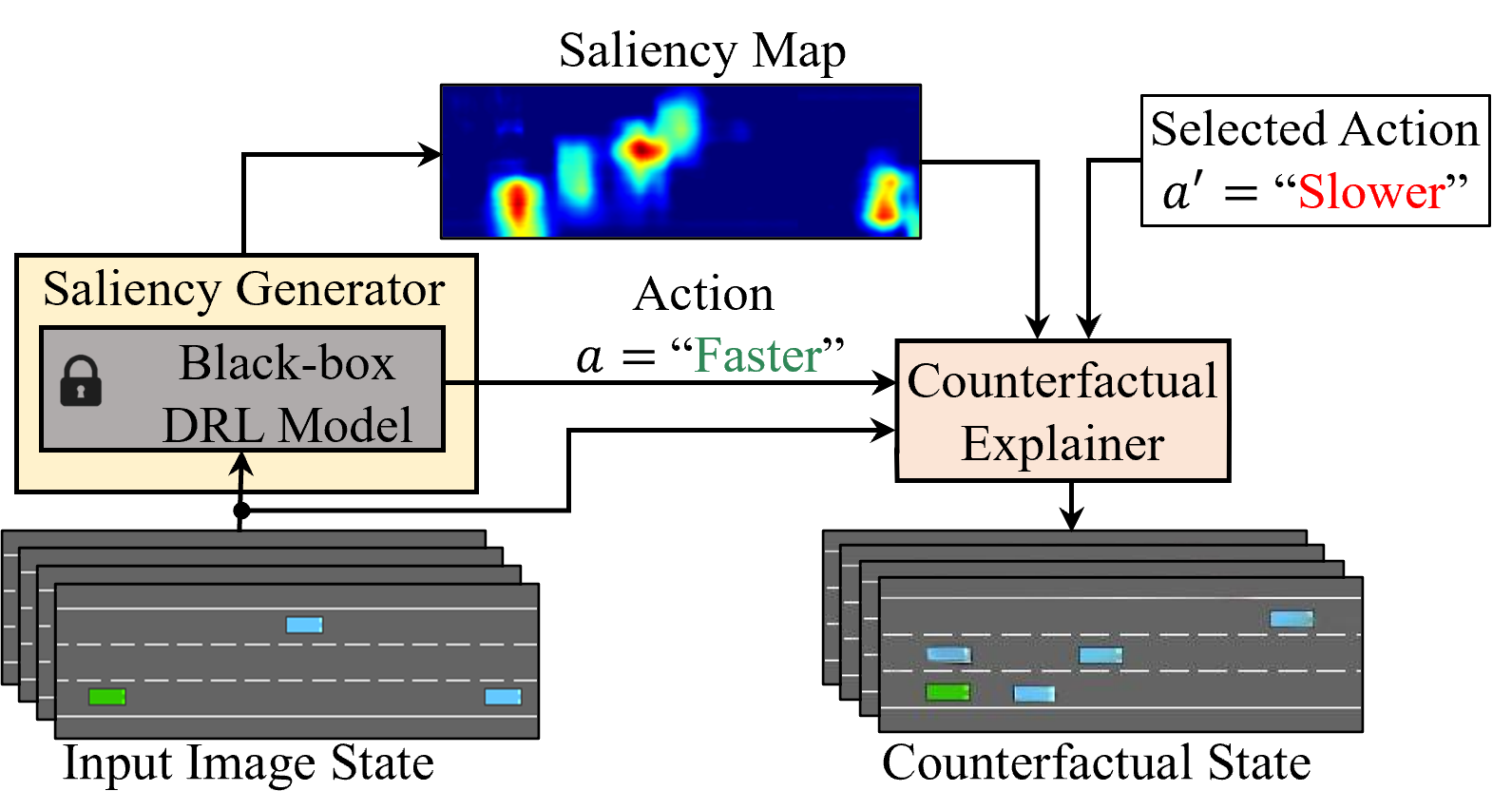}
% \vspace{-0.2cm}
\caption{Overview of the CF generation method for DRL agents. In this illustration, the input and output to the CF explainer consist of four images but only the most recent one is depicted for clarity. While initially, the DRL agent (lock symbolises frozen parameters) decides to accelerate the EGO vehicle (green rectangle) depicted on the left-hand side, the selected action by a user is to slow down the EGO vehicle without performing a lane change. Therefore the SAFE-RL generated a CF state where the EGO vehicle is surrounded by participant vehicles (blue rectangles), depicted on the right-hand side.}
\label{fig: teaser}
\vspace{-0.4cm}
\end{figure}

Several optimisation methods and deep generative models have been recently utilised for generating CF explanations. Optimisation techniques, including perturbation-based, gradient-based, and genetic algorithms, rely on directly modifying the input data or specific features~\cite{samadi2023counterfactual}. Despite demonstrating promising performances with low-dimensional inputs, such as tabular data, they yield incongruous or adversarial examples when applied to high-dimensional inputs, such as images, which have been extensively studied in the context of adversarial attacks~\cite{goodfellow2014explaining}. Deep generative models, such as Generative Adversarial Network (GAN)~\cite{jacob2022steex, zemni2023octet} and Diffusion Models \cite{jeanneret2022diffusion},  overcome this limitation by synthesising CF inputs instead of altering features in the input. However, these models have not been well-investigated in the context of DRL. %Unfortunately, DRL agents operating with visual states pose extra complexity in CF generation because deep generative models should  process multiple state observations at a time. 

%The ability to make sequential decisions in 
DRL agents, in many applications, process a history of observations at every time step resulting in temporal dependencies in the input that add to the complexity of deep generative models creating CF examples~\cite{huber2023ganterfactual}. For instance, in Fig.~\ref{fig: teaser}, a DRL-based ADS receives four state observations of the driving environment in each time step (including the current state and the three most recent past states) before taking action. As a result of this dependency, the CF model should generate four high-resolution images (a CF image per state), which extends its complexity. Besides, the generated CF states 
% at each time step, 
(e.g., the four CF states in Fig.~\ref{fig: teaser}) 
must demonstrate internal consistency, plausibility and a rationale for each complementary action. For instance, in a highway driving scenario, the locations of vehicles in subsequent CF images must represent realistic driving manoeuvres and comply with the speed limits and road geometry. Due to such challenges, to the best of our knowledge, there are merely two available models in the literature~\cite{olson2021counterfactual, huber2023ganterfactual} that generate CF states for DRL agents operating with visual data which are discussed next. 

\subsection{\DIFadd{State-of-the-art}}
\label{sec: SOTA}
To provide explicability for DNNs operating with visual inputs, deep generative models such as GANs have been used to create CF examples, and they are commonly referred to as deep generative CF explainers~\cite{AugustinBC022,jacob2022steex,khorram2022cycle}. One prominent and complicated instance of such models that leverages the latent space of a DRL agent to generate CF images has been proposed by Olson {\it{et al.}} in~\cite{olson2021counterfactual}. That model relies on a complex combination of DNNs, where a DNN generator model is trained to change the agent's action. However, the user cannot specify the desirable alternate action of the DRL agent, which limits the model's applicability. 

 GANterfactual-RL is another deep generative CF model that employs starGAN, an image-to-image (I2I) translation model, to provide CF states~\cite{huber2023ganterfactual}. The objective in I2I translation is to transform an image from one domain to another, e.g., horse to zebra images~\cite{choi2018stargan}. Hence, Huber~{\it{et al.}} in~\cite{huber2023ganterfactual} surmised that generating CF explanations resembles a transformation of the input between different domains. \DIFadd{Nevertheless, the dataset collection algorithm introduced by Huber~{\it et al.}~\cite{huber2023ganterfactual} only saved RGB observation states from the current time-step. Consequently, their GAN model could only generate CF states for a single time-step, which were then converted to grayscale and repeated to cover multiple time-steps. This approach risks misleading labels for the DRL agent, affecting the validity of the method, especially in training the discriminator.} 
 
% To overcome this, our proposed dataset collection now includes grayscale observation states from the last four time steps, alongside RGB data for the current time-step. This enhancement enables our deep generator to provide temporal grayscale states in four channels (each representing a previous time step), along with the RGB state for the current time-step. This improvement enhances clarity and precision for end-users and allows for a more accurate evaluation of validity metrics.} 

Even though both models by Huber~{\it{et~al.}} and Olson~{\it{et~al.}} shed some light on the behaviour of the DRL agents, they often generate implausible or meaningless CFs. This limitation stems from the fact that the applied deep generative models only rely on the state observations without any guidance on how to narrow down the solution space, potentially leading to suboptimal CF solutions in the large visual space domain. Consequently, the existing literature lacks an efficient model capable of generating plausible and meaningful CF examples for DRL agents processing visual states. 

\subsection{\DIFadd{Contributions}}
\label{sec: contribution}
% \textbf{Contributions:}
To bridge the identified gap this paper introduces a novel framework that employs a guided GAN model with saliency maps to produce CF explanations for DRL agents. Saliency maps highlight the most influential input pixels during decision-making, providing valuable insights into the focal regions that capture the most attention from the black-box model. 
% Adapting saliency maps into the DRL domain does not come without cost or challenges. 
In the context of generating CF explanations for supervised learners, we demonstrated, as a preliminary study in~\cite{samadi20203SAFE}, the superiority of incorporating saliency maps in CF generation by restricting the solution space compared to alternative I2I translation models. In this paper, we reformulate and extend our previous model named after SAFE in~\cite{samadi20203SAFE} for use with a DRL agent explainer.

To provide CF explanations for DRL agents, we introduce SAFE-RL, which encompasses substantial enhancements to the generator network to process a sequence of observation states instead of a single image processed in SAFE. This enhancement coupled with revised network layers and loss functions for the generator, have endowed SAFE-RL with spatiotemporal processing capabilities that SAFE lacks. Furthermore, SAFE-RL alleviates the need for the classification objective from its discriminator network, leading to more efficient training and a significant reduction of network parameters by 43\%, as compared to SAFE. Instead, it employs the black-box DRL model for real/fake image classification and introduces a novel loss function in the discriminator. Moreover, SAFE-RL introduces a novel prediction loss for saliency map generation. This change overcomes a limitation in the SAFE model, where the generator simply echoe\DIFadd{d} the input saliency maps \DIFadd{to} generat\DIFadd{e both the} self-attention layer's output (Att) and \DIFadd{the} CF saliency maps. The introduced prediction loss leads the generator network in SAFE-RL toward more accurate saliency map generation compared to SAFE.  % In summary, in this research, we reformulated and extended our previous work in~\cite{samadi20203SAFE} for use with a DRL agent explainer.
%this work improves upon the DNN employed in the deep generative model. Additionally, novel loss functions are proposed to seamlessly integrate the utilisation of saliency maps and to comprehend spatiotemporal state observations.
% we have adjusted the networks used in the SAFE model and proposed novel loss functions to incorporate the use of saliency maps and comprehend the spatiotemporal state observations. 

To encourage further studies in this field, a benchmark is also established, encompassing extensive performance results. New datasets are introduced across various environments and DRL agents\DIFadd{, that provide saliency maps, temporal states' images and DRL agent actions. Particularly, to overcome the aforementioned limitation by Huber  {\it{et al.}}, the proposed dataset collects grayscale observation states from the last four time steps, alongside RGB data for the current time-step. This enhancement enables our deep generator model to provide temporal grayscale states in four channels (each representing a previous time step), along with the RGB state for the current time-step. This improvement enhances clarity and precision for end-users and allows for a more accurate evaluation of validity metrics.} In summary, the key contributions of this paper are:    
\begin{enumerate}
    \item 
    The proposition of SAFE-RL which can be used to interpret DRL agents decisions by integrating a saliency guidance into a GAN for CF generation. 
    %\item Introducing of a novel formulation to fuse the saliency guidance into the DRL-designed GAN, termed SAFE-RL, resulting in the generation of sparse and high-quality CF examples.
    \item A comprehensive performance evaluation analysis of SAFE-RL in three environments for three DRL agents and demonstration of superior performance in terms of validity, sparsity, and proximity for the generated CF examples as compared to baseline methods by~\cite{olson2021counterfactual} and~\cite{huber2023ganterfactual}. We conduct {\color{blue}{45}} experiments in total, which can be considered as a benchmark for future studies. The codes for all experiments are made publicly available. 
    \item The generation of new datasets for multiple environments, including Atari games~\cite{1606.01540}, highway and roundabout driving~\cite{highway-env}, specifically labelled by three popular DRL agents including Deep Q-network (DQN)~\cite{mnih2015human}, Asynchronous Actor-Critic (A2C)~\cite{mnih2016asynchronous} and Proximal Policy Optimisation (PPO)~\cite{schulman2017proximal}. These datasets are made publicly available. %facilitating comprehensive performance analysis and comparisons with baseline methods.
\end{enumerate}

Through the above, this paper aspires to improve the explicability of the decision-making process for DRL agents, advance the field of XAI and facilitate the deployment of transparent and trustworthy AI systems in real-world applications, such as ADS.

%% file: tex/method.tex
This section introduces the proposed SAFE-RL.
% , which relies on saliency maps to guide a deep generative model toward providing CF examples for DRL agents. 
First, we formulate the underlying generation of CF examples for DRL agents. This is followed by the details of SAFE-RL and its training procedure. 

\subsection{Problem Definition}
\label{subsec: problem_def}
RL agents are commonly utilised within a Markov Decision Process~(MDP), a mathematical framework that encompasses a set of states $s \in S$, a set of actions $a \in A$ and rewards $R(s,a)$. The objective of the agent is to find a policy $\pi(a|s)$ that maximises the expected cumulative reward $E[\sum ^{\infty}_{t=0}\gamma^t R(s_t,a_t)]$ discounted with a time-dependent non-negative factor $\gamma^t$ controlling the importance of future rewards as compared to the immediate reward $(\gamma \leq 1)$. The policy function is a probabilistic mapping of states to actions  $\pi:S \rightarrow A$, and the state-action pair at the $t$-th time step has been denoted by $(s_t,a_t)$. In the realm of DRL, agents commonly receive a historical batch containing a series of $h$ sequential past state observations. To represent this sequence of subsequent states, we denote a state as $\mathbf{s}$, where $\mathbf{s}_t = \{ s_{t-j}\}_{j=0}^{h-1}$.

To interpret the behaviour of DRL agents and explain why a particular action $a$ is selected at a state $\mathbf{s}$, CF reasoning exemplifies a modified state $\mathbf{s}'$ that results in a complement action, such that $\pi(\mathbf{s}') = a'$,  where $a'\neq a$. Meanwhile, minimising the alternations (i.e. keeping the CF state $\mathbf{s}'$ as close as possible to the original state $\mathbf{s}$) provides insights into the
rationale behind the selection of the action $a$, by highlighting the necessary modifications that would lead to a different action.
The problem addressed in this paper is to provide such CF examples for DRL agents by using a saliency-aware deep generative model.

\subsection{Overview of the Proposed Framework}
\label{subsec: SAFE_overview}
This subsection provides an overview of the SAFE-RL including the generation of the training dataset and the CF generation process. Since the generation of the saliency maps requires the weights of the DRL agent, we hereafter referred to it as grey-box DRL agent.  

% This subsection provides an overview of the SAFE-RL approach, which utilizes saliency maps from a given DRL agent to generate CF examples. The procedure involves dataset generation and the CF generation process using the SAFE-RL framework.

\subsubsection{Dataset}
SAFE-RL relies on a training dataset $\mathcal D$ that comprises state observations paired with actions and saliency maps, see Fig. 2. The $i$-th element of the dataset is denoted by $\mathcal{D}_i = {({\mathbf{s}}_i, m_i, a_i)}^N_{i=1}$, where \DIFadd{${\mathbf{s}}_i=(s_i^{rgb}, s_i^{grey}, s_{i-1}^{grey}, s_{i-2}^{grey}, s_{i-3}^{grey})$} is the vector of the current \DIFadd{(in both rgb and grey-sacle)} and three past observation states and $a_i$ is the selected action at the $i$-th time step by the DRL agent. \DIFadd{Notably, grayscale state images contribute more to the training process, while the RGB state image enhances understanding and traceability for end-users}. 
%This information is collected from a given grey-box DRL operating in an environment (see Fig. \ref{fig: dataset}). 
%In this setup, within a running MDP, we save temporal state observations, for instance, history stacks of four, represented as $[s_i^{t-j}, (j=0,1,2,3)]$. 
%Simultaneously, applying the Eigen-CAM method \cite{muhammad2020eigen} to the DRL, we obtain the saliency map indicating the contribution of input pixels (state) in the latest state frame to the DRL agent's output (action). 
For a state image vector ${\mathbf{s}}_i \in S$, we use the Eigen-CAM method~\cite{muhammad2020eigen} to generate the saliency map $m_i = M({\mathbf{s}}_i)$, which is a 2D matrix highlighting the salient pixels of the input images for the DRL agent. The dimensions of $m_i$ are equal to that of an input image $s_i$ and its elements take values in the range $(0,1)$. 
% The dataset element  $\mathcal{D}_i$ is stored in the Replay Memory, see Fig.~\ref{fig: dataset}.
We have applied SAFE-RL to three grey-box DRL agents, including PPO, DQN and A2C, operating in the highway, roundabout and Atari Pong game environments. Each one of the nine generated datasets has been divided into training, validation and test sets with proportions of 80~\%, 10~\% and 10~\%, respectively. 
% To start, our approach is built upon a training dataset denoted as $\mathcal{D}$, which consists of state observations paired with their corresponding saliency maps and actions. Specifically, $\mathcal{D} = {(s_i, m_i, a_i)}_{i=1}^N$, where $s_i$ represents state observations, $m_i$ is the corresponding saliency map, and $a_i$ denotes the action taken in response to the state. This data is collected from a grey-box Deep Reinforcement Learning (DRL) system operating within a specific environment (see Fig. \ref{fig: dataset}). In our setup, as the Markov Decision Process (MDP) runs, we store temporal state observations. For example, we save history stacks of the last four observations as $[s_i^{t-j}, (j=0,1,2,3)]$. Simultaneously, we utilize the Eigen-CAM method \cite{muhammad2020eigen} on the DRL system to generate saliency maps. These maps highlight the input pixels (state) that contribute the most to the DRL agent's output (action). Given an input state image $s \in S$ paired with action $a \in A$, the saliency map generator returns a 2D map $m = M(s, a)$, indicating the salient regions of the input image concerning the DRL agent's output. Subsequently, the DRL agent responds to the given state by providing an action, which is then stored in the Replay Memory. To evaluate and test our approach, we divide the gathered dataset into training and test sets. For further details and access to the datasets, please refer to the project link.
\begin{figure}[t]
\centering
\includegraphics[width=0.88\linewidth, height=4.1cm]{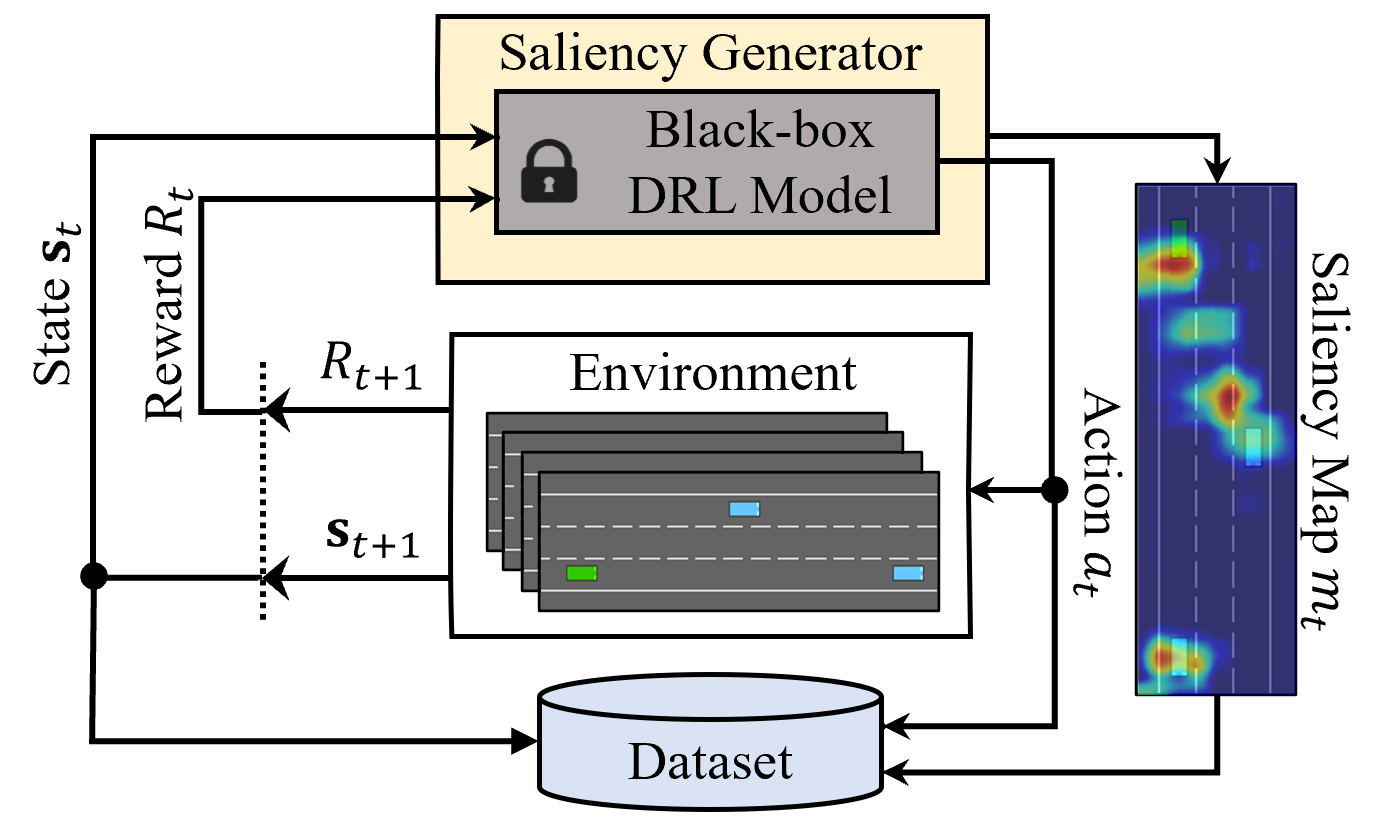}
\vspace{-0.1cm}
\caption{Generation process of the dataset including the states, actions and saliency maps for a grey-box DRL agent.} 
% In DRL terminology, the dataset is commonly referred to as replay memory.}
\label{fig: dataset}
\vspace{-0.2cm}
\end{figure}

\subsubsection{SAFE-RL}
\begin{figure*}
\centering
\hspace{-1cm}
\begin{subfigure}{0.64\linewidth}
\centering
\includegraphics[height=4.8cm]{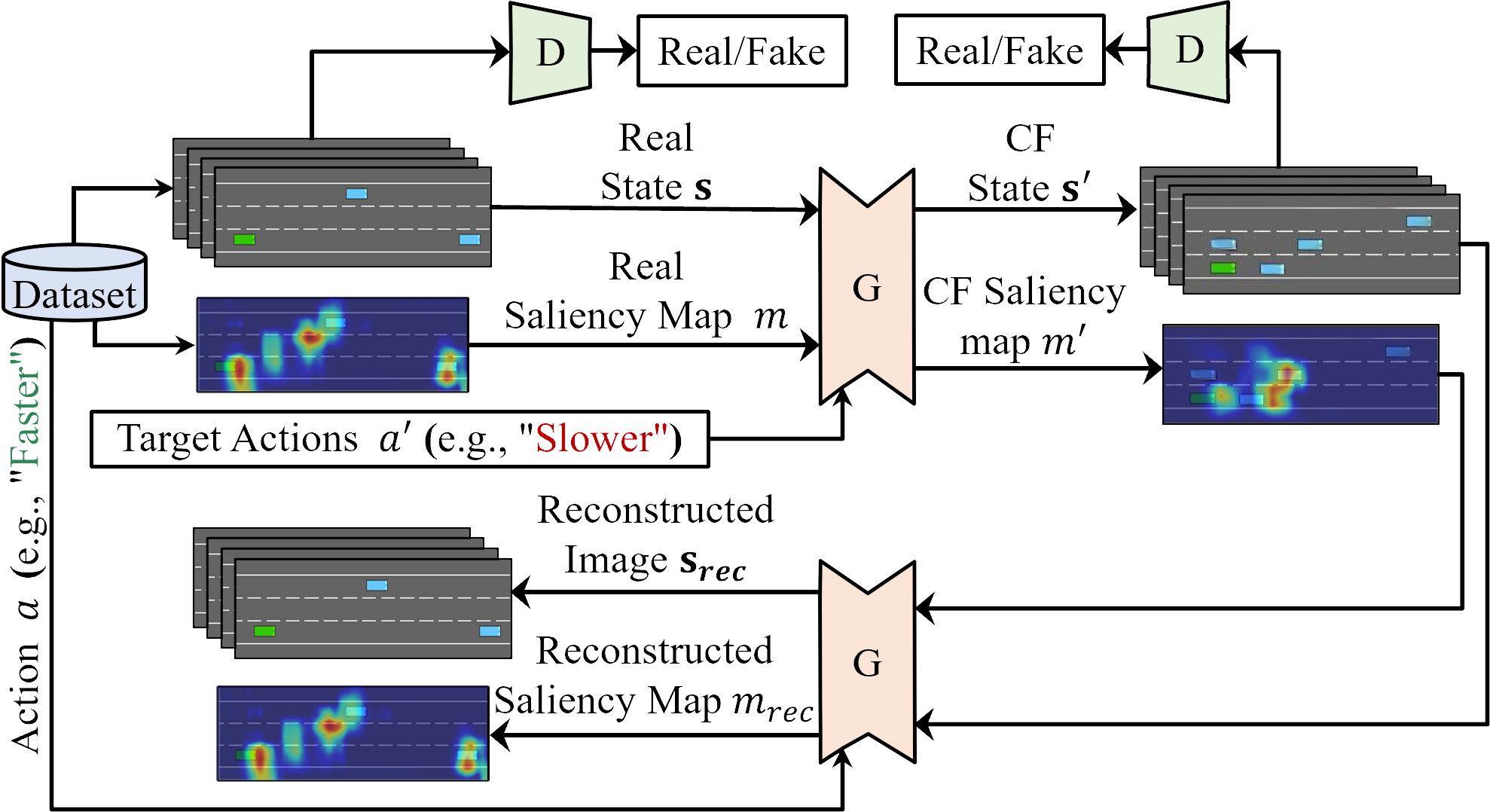}\\{\vspace{-0.05cm}\scriptsize (a)}
\end{subfigure}
% \hspace{.3cm}
\begin{subfigure}{0.35\linewidth}
\centering
\hspace{-0.5cm}
\includegraphics[height=3.cm]{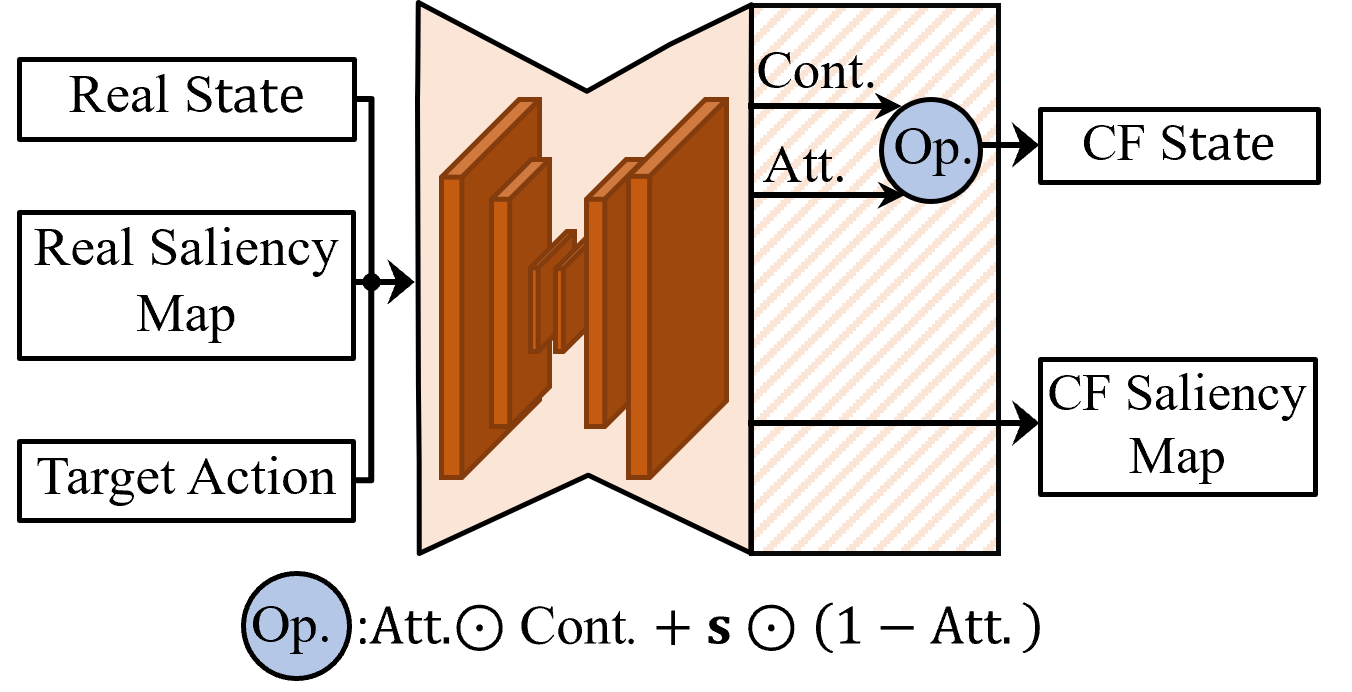}
\\{\vspace{-0.05cm}\scriptsize (b)}
\end{subfigure}
\caption{(a) Block diagram of the SAFE-RL model. (b) Generator network model details.}
\label{fig: method0}
\vspace{-0.3cm}
\end{figure*}

%The primary objective of CF generation is to apply minimal modifications to the original input/state, resulting in an alteration in the output/action. To achieve this, the deep generative model is guided by saliency maps to apply alterations specifically to these salient regions. 
Fig.~\ref{fig: method0} illustrates the training process of the SAFE-RL.
At each time step, the generator $G(\cdot)$ is fed with the vector of states ${\mathbf{s}}_i$, the saliency map $m_i$, and the counter action $a_i'\neq a_i$. Its goal is to generate CF states ${\mathbf{s}}_i'$ that can be associated with the action $a_i'$ such that $a_i' = \pi(\mathbf{s}_i')$. To ensure the generation of realistic and plausible CF states, the generator is coupled with the discriminator $D(\cdot)$, which is trained to distinguish between CF and real states. Through an adversarial dynamic, the generator is led to provide CF states that follow the real states distribution such that the discriminator classifies the generated states as real ones. In the SAFE-RL framework (Fig.~\ref{fig: method0}(a)), the generator model also transfers states from the CF domain to the original domain, ensuring consistency in domain transformations. 

In this study, we have employed attentionGAN~\cite{tang2021attentiongan} as the backbone architecture of our generator network. 
This architecture provides the CF/fake states from a combination of a generative image, known as the content (represented as 'Cont.' in Fig.~\ref{fig: method0}b) and the real states $\mathbf{s}$ by using a self-attention layer ('Att') as a pixel selector. This entails a two-stage generation process where in the attended regions, we incorporate the pixels from the content layer, while in non-attended regions, we assimilate pixels from the real state, see the operation (Op.) in Fig.~\ref{fig: method0}b. This strategy has been demonstrated to yield sharper and more realistic image translation outcomes as compared to the starGAN~\cite{choi2018stargan}. 
Note that the saliency map in SAFE-RL indicates the significant pixels within the states for the DRL model, while the 'Att' layer signifies the attended regions for the generator. Intuitively, the saliency map can serve as a guiding mechanism for the generator's output. In the subsequent paragraphs, we outline the employed loss functions that contribute to achieving the goal of generating realistic CF states.

To begin with, the loss functions set the adversarial dynamic, where the discriminator and generator improve their performance by minimising the following adversarial loss: %$\mathcal{L}^G_{{adv}}$ and $\mathcal{L}^D_{{adv}}$ 
%that drive the operation of the generator and discriminator respectively to improve their performances and can be read as:
% \vspace{0.15mm}
% \begin{figure}
%     \centering
%     \begin{subfigure}{\linewidth}
%         \centering
%         \includegraphics[width=1\linewidth]{img/model.png}\\{\vspace{-0.15cm}\scriptsize (a) Block diagram of the SAFE-RL model.}
%     \end{subfigure}
% % \hspace{0.2cm}
%     \begin{subfigure}{\linewidth}
%         \centering
%         \includegraphics[width=0.8\linewidth]{img/generator2.png}\\{\vspace{-0.15cm}\scriptsize (b) Generator network model details.}
%     \end{subfigure}
% \caption{Overview of the SAFE-RL}
% \label{fig: method0}
% % \vspace{-0.4cm}
% \end{figure}

\begin{equation}
\label{eq: loss_adv}
\begin{gathered}
    \mathcal{L}_{D}^{adv} = -\mathbb{E}_{\mathbf{s}}\left[D(\mathbf{s})\right] + \mathbb{E}_{\mathbf{s},a'}
    \left[D(\mathbf{s}') \right] - \lambda_{gp}\mathcal{L}^{gp}_D,
    % \lambda_{gp} \, \mathbb{E}_{\hat{s}}\Bigr[(\lVert \nabla_{\hat{s}} D_{src}(\hat{s}) \rVert_2-1 )^2\Bigr],\\
    % \hat{s} = \alpha  s + (1 - \alpha) s', 
\end{gathered}
\end{equation}
% \begin{equation}
% \label{eq: x_hat}
% \end{equation}
\begin{equation}
    \mathcal{L}_G^{adv} = - \mathbb{E}_{\mathbf{s},a'} \left[D(\mathbf{s}') \right],
\end{equation}
where  $\lambda_{gp}>0$ is a hyperparameter controlling the contribution of the gradient penalty $\mathcal{L}_{gp}^D$ to the loss function, and the expectations $\mathbb{E}_{\mathbf{s},a'}[\cdot]$ are taken over a batch of input states $\mathbf{s}$ and all possible complementary actions $a'\neq a$. 
The gradient penalty term promotes smooth gradients along the data manifold, stabilises the training process and mitigates the mode collapse issue~\cite{gulrajani2017improved}. To this end, it ensures that the discriminator's gradient remains within a reasonable range (between negative and positive unity) by minimising a parabolic function of its norm at a randomly sampled point $\hat{\mathbf{s}}$ lying between an original sample $\mathbf{s}$ and a generated CF sample $\mathbf{s}'$:
\begin{equation}
\mathcal{L}^{gp}_D = \mathbb{E}_{\hat{\mathbf{s}}}\left[\left(\lVert \nabla D(\hat{\mathbf{s}}) \rVert_2-1 \right)^2\right],
\end{equation}
where 
% \begin{equation}
$\hat{\mathbf{s}} = \alpha \mathbf{s} + (1 - \alpha) \mathbf{s}',$
% \end{equation}
with $\alpha$ being a uniformly selected random number in the range $(0, 1)$, and $\lVert \nabla D(\hat{\mathbf{s}}) \rVert_2$ is the Euclidean norm of the discriminator's gradient evaluated at the point $\hat{s}$.

While the loss function of the discriminator $\mathcal{L}_D$ is equal to the adversarial loss $\mathcal{L}_D=\mathcal{L}_D^{adv}$ in order to meet the objective of distinguishing real and CF states, the generator is subject to multiple loss functions in order to fulfill its objectives, i.e., producing realistic CF states by effectively incorporating saliency map details. These distinct loss functions are elaborated upon in the ensuing discussion.

To ensure that the generated CF states $\mathbf{s}'$ are mapped to the desired action $a'$ by the DRL agent, we include a term in the generator's loss function, which minimises the cross entropy between the desired action and the output of the DRL model in response to the CF states. Due to the fact that the desired action is not a random variable, this loss term, denoted by $\mathcal{L}_{G}^{{cls}}$, takes the following simplified form 
%the $\mathcal{L}^{G}_{{cls}}$  
% The following , leads the generator network towards generating CF states $\mathbf{s}'$ such that the DRL model classifies them as desired actions $a'$, i.e. $\pi(\mathbf{s}')=a'$:
% \begin{equation}
% \begin{gathered}
%     \mathcal{L}^y_{{cls}} = \mathbb{E}_{s,a} \Bigr[-\log D_{cls}(a|s)\Bigr],
% \end{gathered}
% \end{equation}
\begin{equation}
    \mathcal{L}^{cls}_{G} = \mathbb{E}_{\mathbf{s},a'}\left[-\text{log}\pi(a'|\mathbf{s}')\right],
\end{equation}
where the expectation is calculated across a batch of input states $\mathbf{s}$ and over all complementary actions $a'\neq \pi(\mathbf{s})$. % which are associated with the CF states $\mathbf{s}'$.
% where $\pi(s')$ stands for DRL agent decision, that ideally should be $a'$. 

% Considering the fact that $\mathcal{L}^{G}_{{cls}}$ will be used in generator objectives (see Eq. \ref{eq: L_G}), the usage of $\pi(s')$ could help in conveying consistent classification expectation in the generator, since  with expecting a in discriminator to learn the generator.

Furthermore, to ensure consistency, the generated CF states (fake images) are transferred back into the initial domain (see Fig.~\ref{fig: method0}) and compared with the query states. The reconstruction loss, $\mathcal{L}_{G}^{rec}$, penalises the generator for adding excessive artifacts and enforces forward and backward consistency. Specifically, one may write 
\begin{equation}
%\binom{x'}{s'} = G_{Y' \to Y}(x, s,y').\\
\mathcal{L}_{G}^{rec} = \mathbb{E}_{\mathbf{s},a'} \Bigr[ \big\|{ (\mathbf{s},m) -  G(G(\mathbf{s}, m,a'),a)}  \big\|_1 \Bigr], 
\end{equation}
where the output of the generator (forward transfer) can be read as $(\mathbf{s}',m')=G(\mathbf{s}, m,a')$ and a similar expression can be written for the backward transfer. To calculate the vector norm, $\lVert \cdot \rVert_1$, the pairs of states $\mathbf{s}$ and the saliency map $m$ should be concatenated into a column vector. % that is described by the transpose operation $(\cdot,\cdot)^T$.

% In \cite{samadi20203SAFE} 
Another term in the generator's loss function is introduced next to ensure the effective integration of saliency map information into the generator network. The role of this term that is denoted by $\mathcal{L}_G^{fuse}$ and referred to as the fuse loss function is to impose a penalty on the generator for altering non-salient features. %, ensuring that the model perceives and alters only the salient region's pixels to provide CF examples. 
\begin{equation}
\label{eq:lfuse}
    \mathcal{L}_{G}^{fuse} = \mathbb{E}_{\mathbf{s},a'}\Bigr[ || (\mathbf{s}-\mathbf{s}')\odot (1-m) ||_1\Bigr],
\end{equation}
where $\odot$ denotes element-wise multiplication of 2D matrices. Note that the complementary of the saliency map, $(1-m)$, has to be replicated four times because, in our case, the input state $\mathbf{s}$ consists of four images. The result of the point-wise multiplication has to be converted into a column vector before the vector norm $\lVert \cdot \rVert_1$ can be finally computed.

The last term of the loss function 
%However, we have identified that this approach does not guarantee the accurate generation of genuine saliency maps for CF state images. 
%It has been observed that the generator is able to use those information and apply changes into the salient regions, and propagate the same real saliency map as fake saliency maps, leading to a zero loss in the $\mathcal{L}_{rec}$ loss component, which pertains to saliency map reconstruction. As a remedy, we 
introduces the saliency prediction loss, $\mathcal{L}_G^{pred}$, which involves a comparison between the CF saliency map $m'$ and the map generated using the Eigen-CAM method $M(\mathbf{s}')$. One may write
\begin{equation}
\label{eq:lfuse}
    \mathcal{L}_{G}^{pred} = \mathbb{E}_{\mathbf{s},a'}\Bigr[ || m' -M(\mathbf{s}')||_1\Bigr].
\end{equation}
\DIFadd{By leveraging the ground-truth CF saliency map, $M(\mathbf{s}')$, this loss term mitigates the limitation of naively replicating the real saliency map as CF saliency map and ``att'' layer output, as has been mentioned in Section \ref{sec: contribution}. Essentially, it guides the generator in understanding how the DRL model allocates attention to the input state pixels, thereby enhancing its ability to generate CF states proficiently.}

At last, a linear combination of the presented loss functions defines the objective of the generator model as follows: % denoted as $\mathcal{L_\text{G}}$.
% \begin{equation}
% \label{eq: L_D}
%     \mathcal{L}_D = -\mathcal{L}_{adv}^D.
% \end{equation}
\begin{equation}
\label{eq: L_G}
    {\mathcal{L}_{G}} = -\mathcal{L}_G^{adv} + \lambda_{cls} \mathcal{L}^{cls}_{G} + \lambda_{{fuse}} \mathcal{L}_{G}^{fuse}+\DIFadd{\lambda_{{pred}}}\mathcal{L}_{G}^{pred},
\end{equation}
where the parameters $\lambda_{cls}$, $\lambda_{fuse}$ \DIFadd{and $\lambda_{pred}$} are positive scalars.

% \subsection{Algorithm}
% % \vspace{-0.1cm}

% Algorithm~\ref{alg: alg} provides a pseudocode summary of the training procedure for the SAFE-RL model, while the specific loss functions are detailed in the main paper. 
% \vspace{-0.4cm}

\RestyleAlgo{ruled}

%% This is needed if you want to add comments in
%% your algorithm with \Comment
\SetKwComment{Comment}{\textcolor{teal}{/* }}{ \textcolor{teal}{/* }}
\SetKwInput{kwReturn}{return}

%% file: tex/results.tex
This section conducts an extensive evaluation of the proposed SAFE-RL framework against the baseline methods~\cite{olson2021counterfactual, huber2023ganterfactual} for three grey-box DRL agents: DQN~\cite{mnih2015human}, A2C~\cite{mnih2016asynchronous} and PPO~\cite{schulman2017proximal}. 
%These analyses are carried out in two applications: 1) automated/autonomous driving in highway and roundabout driving scenarios \cite{highway-env} and 2) Atari Pong game \cite{1606.01540}. 
We first present the implementation details of SAFE-RL followed by the metrics used to evaluate the quality of the generated CF explanations and the selected environments to conduct our experiments. This section culminates with the report of the performance evaluation results and the discussion of several qualitative scenarios that illustrate the effectiveness of the generated CF examples in conveying human-understandable explanations to the end-users.

\definecolor{grey}{rgb}{0.9,0.9,0.9}
\begin{table*}
\begin{center}
\caption{Quantitative comparison of SAFE-RL.{\small{**}}}
\label{tab: performance}
% \vspace{-0.2cm}
\begin{adjustbox}{width=\linewidth,, height=1.9cm,center}
\begin{tblr}{
    column{3-22} = {c},
  cell{1}{1} = {r=2}{l},
  cell{1}{2} = {r=2}{l},
  cell{1}{3} = {c=5}{c},
  cell{1}{8} = {c=5}{c},
  cell{1}{13} = {c=5}{c},
  cell{1}{18} = {c=5}{c},
  cell{1}{23} = {c=5}{c},
  cell{3}{1} = {r=3}{},
  cell{6}{1} = {r=3}{},
  cell{9}{1} = {r=3}{},
  hline{1} = {-}{1pt},
  hline{2} = {4-6,9-11,14-16,19-21,24-26}{0.6pt},
  hline{2} = {3,8,13,18,23}{l,0.6pt},
  hline{2} = {7,12,17,22,27}{0.6pt},
  % hline{2} = {8-11}{c},
  % hline{2} = {11}{r},
  % hline{2} = {12-15}{c},
  % hline{2} = {4-8}
  % hline{2} = {9-13}
  % hline{2} = {13-17}
  % hline{2} = {17-21}
  % hline{2} = {9-13,21-25}{},
  hline{3} = {-}{},
  % hline{3} = {1-8,10-15,17-22}{},
  % hline{3} = {8-10,15-17}{},
  % % hline{6,9} = {1-8,10-15,17-22}{},
  hline{12} = {-}{1pt},
  % colspec = {l X[c] X[c] X[c] X[c] X[c] X[c] X[c] X[c] X[c] X[c] X[c] X[c] X[c] X[c] X[c] X[c] X[c] X[c] X[c]},
  column{1} = {l}{0.45cm},
  column{2} = {c}{0.59cm},
  column{3-20} = {c}{0.5cm},
  column{3-5, 8-10, 13-15, 18-20, 23-25} = {c}{0.47cm},
  column{25} = {c}{0.57cm},
  column{6, 11, 16, 21, 26} = {c}{0.47cm},
  % column{8,14,20,26} = {c}{0.0cm},
  row{3-5,9-11} = {grey},
  % row{11} = {m}{0.3cm},
  % row{10} = {m}{0.3cm},
  row{-} = {m}{0.1cm},
  % column{4,10,16} = {c}{1cm},
}
DRL & Method & Highway & & & & & Roundabout & & & & & Pong & & & & & \DIFadd{SpaceInvaders} & & & & & \DIFadd{Mspacman} & & & &\\
&&  $\downarrow$FID & $\downarrow$LPS & $\downarrow$Prx & $\downarrow$Spr{\scriptsize\%} & $\uparrow$Vld{\scriptsize\%} &
    $\downarrow$FID & $\downarrow$LPS & $\downarrow$Prx & $\downarrow$Spr{\scriptsize\%} & $\uparrow$Vld{\scriptsize\%} &
    $\downarrow$FID & $\downarrow$LPS & $\downarrow$Prx & $\downarrow$Spr{\scriptsize\%} & $\uparrow$Vld{\scriptsize\%} &
    \DIFadd{$\downarrow$FID} & \DIFadd{$\downarrow$LPS} & \DIFadd{$\downarrow$Prx} & \DIFadd{$\downarrow$Spr{\scriptsize\%}} & \DIFadd{$\uparrow$Vld{\scriptsize\%}} &
    \DIFadd{$\downarrow$FID} & \DIFadd{$\downarrow$LPS} & \DIFadd{$\downarrow$Prx} & \DIFadd{$\downarrow$Spr{\scriptsize\%}} & \DIFadd{$\uparrow$Vld{\scriptsize\%}} \\

DQN&Olson   &2.049          &0.405          &12.99          &99.27          &\ul{99.82}     &NC            &NC             &NC             &NC             &NC             &\textbf{0.178}&0.163         &4.369          &95.58          &\ul{91.81}			&\DIFadd{\ul{0.445}}     &\DIFadd{0.301}         &\DIFadd{8.113}         &\DIFadd{53.11}         &\DIFadd{\textbf{100}}  &\DIFadd{2.449}             &\DIFadd{0.120}             &\DIFadd{16.410}        &\DIFadd{90.16}         &\DIFadd{\textbf{100}}	\\
    &Huber  &\ul{1.353}     &\ul{0.127}     &\ul{2.739}     &\ul{79.50}     &73.88          &3.505         &0.150          &\textbf{0.014} &100            &81.65          &\ul{0.181}    &\ul{0.087}    &\ul{1.012}     &\ul{56.80}     &66.25					&\DIFadd{4.683}          &\DIFadd{\ul{0.197}}    &\DIFadd{\ul{1.232}}    &\DIFadd{\ul{23.88}}    &\DIFadd{29.60}         &\DIFadd{\ul{2.238}}        &\DIFadd{\ul{0.114}}        &\DIFadd{\ul{4.156}}    &\DIFadd{\ul{73.57}}    &\DIFadd{37.01}			\\
    &\textbf{Ours} &\textbf{1.282} &\textbf{0.049} &\textbf{1.559} &\textbf{57.90} &\textbf{100}   &\textbf{0.621}&\textbf{0.017} &0.892          &\textbf{48.47} &\textbf{100}   &0.225         &\textbf{0.064}&\textbf{0.862} &\textbf{55.90} &\textbf{94.27}	&\DIFadd{\textbf{0.064}}	&\DIFadd{\textbf{0.001}}&\DIFadd{\textbf{0.816}}&\DIFadd{\textbf{19.81}}&\DIFadd{\ul{45.97}}    &\DIFadd{\textbf{0.188}}    &\DIFadd{\textbf{0.005}}    &\DIFadd{\textbf{1.680}}&\DIFadd{\textbf{55.69}}&\DIFadd{\ul{41.63}}	\\
PPO&Olson   &\ul{0.866}     &0.159          &\ul{2.331}     &88.60          &\ul{80.59}     &NC            &  NC           & NC            & NC            &NC             &\textbf{0.114}&0.164         &3.978          &91.97          &71.28					&\DIFadd{\textbf{0.180}} &\DIFadd{0.219}         &\DIFadd{5.402}         &\DIFadd{65.58}         &\DIFadd{\textbf{100}}  &\DIFadd{2.273}             &\DIFadd{0.138}             &\DIFadd{16.613}        &\DIFadd{91.09}         &\DIFadd{\textbf{100}}	\\
    &Huber   &1.466          &\ul{0.122}     &2.426          &\ul{73.38}     &75.19          &2.150         &0.103          &\textbf{0.006} &100            &73.14          &\ul{0.130}    &\ul{0.136}    &\ul{1.409}     &\ul{82.33}     &\ul{85.04}			&\DIFadd{\ul{4.991}}     &\DIFadd{\textbf{0.179}}&\DIFadd{\ul{0.929}}    &\DIFadd{\ul{20.21}}    &\DIFadd{35.43}         &\DIFadd{\textbf{1.865}}    &\DIFadd{\ul{0.076}}        &\DIFadd{\ul{3.220}}    &\DIFadd{\ul{72.06}}    &\DIFadd{41.00}			\\
    &\textbf{Ours} &\textbf{1.246} &\textbf{0.023} &\textbf{1.493} &\textbf{52.36} &\textbf{99.97} &\textbf{0.280}&\textbf{0.008} &0.705          &\textbf{44.65} &\textbf{99.93} &0.165         &\textbf{0.053}&\textbf{0.848} &\textbf{56.00} &\textbf{99.61}	&\DIFadd{7.129}          &\DIFadd{\ul{0.209}}    &\DIFadd{\textbf{0.602}}&\DIFadd{\textbf{17.11}}&\DIFadd{\ul{56.82}}    &\DIFadd{\ul{2.012}}             &\DIFadd{\textbf{0.064}}    &\DIFadd{\textbf{1.633}}&\DIFadd{\textbf{55.87}}&\DIFadd{\ul{43.31}}	\\
A2C&Olson   &\textbf{0.088} &0.164          &\ul{2.413}     &89.28          &88.38          &NC            &NC             &NC             &NC             &NC             &5.065         &0.417         &5.560          &92.76          &\ul{95.64}			&\DIFadd{\ul{0.429}}     &\DIFadd{0.284}         &\DIFadd{6.714}         &\DIFadd{71.04}         &\DIFadd{\textbf{70.62}}&\DIFadd{\textbf{1.959}}    &\DIFadd{0.212}             &\DIFadd{18.160}        &\DIFadd{93.29}         &\DIFadd{\textbf{75.72}}\\
   &Huber   &1.447          &\ul{0.130}     &3.636          &\ul{59.25}     &\ul{78.05}     &4.778         &0.362          &\textbf{0.073} &100            &28.08          &\ul{0.266}    &\ul{0.120}    &\ul{1.371}     &\ul{80.09}     &67.87 				&\DIFadd{4.008}          &\DIFadd{\ul{0.166}}    &\DIFadd{\ul{1.043}}    &\DIFadd{\ul{21.62}}    &\DIFadd{34.27}         &\DIFadd{2.419}             &\DIFadd{\ul{0.204}}        &\DIFadd{\ul{5.745}}    &\DIFadd{\ul{73.32}}    &\DIFadd{24.61}			\\
   &\textbf{Ours} &\ul{1.290}     &\textbf{0.072} &\textbf{1.818} &\textbf{48.85} &\textbf{98.97} &\textbf{1.157}&\textbf{0.045} &0.788          &\textbf{48.05} &\textbf{60.00} &\textbf{0.061}&\textbf{0.051}&\textbf{1.064} &\textbf{61.08} &\textbf{100}  	&\DIFadd{\textbf{0.074}} &\DIFadd{\textbf{0.027}}&\DIFadd{\textbf{0.724}}&\DIFadd{\textbf{14.04}}&\DIFadd{\ul{48.42}}    &\DIFadd{\ul{2.215}}        &\DIFadd{\textbf{0.053}}    &\DIFadd{\textbf{1.501}}&\DIFadd{\textbf{57.34}}&\DIFadd{\ul{39.01}}	\\
\end{tblr}
\end{adjustbox}
\end{center}
\vspace{-0.2cm}
\scriptsize{{\small{**}}The desired directions of metrics improvement are indicated by arrows ($\downarrow$ or  $\uparrow$). The most favorable results are highlighted in bold, and the second-bests are underscored. LPIPS, proximity, sparsity, and validity metrics are represented by LPS, Prx, Spr, and Vld, respectively. The methods that were unable to converge are denoted by NC.}
\vspace{-0.2cm}
\end{table*}

\begin{table}
\begin{center}
\caption{\DIFadd{Ablation study over Highway Environment.{\small{**}}}}
\label{tab: abilation}
\vspace{-0.2cm}
{\color{black}
\begin{adjustbox}{width=\linewidth, height=1.1cm ,center}
\begin{tblr}{
  hline{1,6} = {-}{1 pt},
  hline{2} = {-}{},
  row{1-5} = {m}{0.2cm},
  row{1-5} = {m}{0.2cm},
  column{1-7} = {c},
  column{1-2} = {c}{0.6cm},
  column{3-7} = {c}{1.1cm},
}
$\lambda_{fuse}$ & $\lambda_{pred}$ & $\downarrow$FID & $\downarrow$LPIPS & $\downarrow$Prx & $\downarrow$Sprs{\scriptsize\%} & $\uparrow$Vld{\scriptsize\%}\\
    0   &0  &1.392  &0.113  &2.274  &78.23  &77.43\\
    0   &1  &1.504  &0.115  &2.061  &77.05  &\ul{94.51}\\
    1   &0  &\textbf{1.145}  &\textbf{0.038}  &\textbf{1.460}  &\textbf{55.47}  &81.35\\
    1   &1  &\ul{1.282}  &\ul{0.049}  &\ul{1.559}  &\ul{57.90}  &\textbf{100}  
\end{tblr}
\end{adjustbox}
}
\end{center}
\vspace{-0.2cm}
\DIFadd{\scriptsize{{\small{**}}The desired directions of metrics improvement are indicated by arrows ($\downarrow$ or  $\uparrow$). The best results are highlighted in bold, while the second-best model is underlined. Proximity, sparsity, and validity are denoted by prx, sprs, and vld, respectively.}}
\vspace{-0.2cm}
\end{table}

\subsection{Implementation Details}
\vspace{-1mm}
\label{sec:ID}
In our experiments, we use four stacked states as the observation history size for all environments: Highway and ~\cite{highway-env}, Roundabout, and Atari games~\cite{1606.01540}; the pixel resolution sizes for these environments are $128\times512$, $448\times448$, and $84\times84$, respectively. We repeat each experiment three times with randomly generated seeds and present the mean values of the extracted metrics. We employ the Adam optimiser associated with a learning rate of $10^{-4}$. The hyper-parameter and loss coefficients are determined heuristically and configured as follows: $\lambda_{cls}=1, \lambda_{gp}=10, \lambda_{rec}=10$\DIFadd{,} $\lambda_{fuse}=1$ \DIFadd{and $\lambda_{pred}=1$} for all experiment, see Eq.~(1) and Eq.~(8). The SAFE-RL framework leverages the generator and discriminator networks from the AttentionGAN models~\cite{tang2021attentiongan} with some modifications applied to layers, input/output and loss functions. \DIFadd{For fair evaluation, we have set the same hyperparameters for the Huber~{\it{et al.}}'s model because that model is based on the StarGAN backbone, which is the original backbone of AttentionGAN used by SAFE-RL. For Olson~{\it{et al.}}, we conducted a grid-based search across varying dimensions, including the ``state latent space'' $(Z_s\in[16, 32])$, ``agent latent space'' $(Z_a\in[256, 512])$, and ``Wasserstein auto-encoder latent space'' $(Z_{wae}\in[64, 128])$, ultimately selecting the optimal configuration to yield the best results.} As for the subject grey-box DRL agent, we use pure stable-baselines3 \cite{stable-baselines3} DRL agents, as illustrated in Fig.~\ref{fig: method0}. 
Additionally, we incorporate the Eigen-CAM method \cite{muhammad2020eigen} as the ``Saliency Generator'' component. 
% For the implementation, we utilise the PyTorch framework, harnessing the processing power of an NVIDIA RTX-3090 GPU.

\subsection{Performance Metrics}
\label{sec: metric}
To evaluate the quality of the generated CF examples, we employ well-established CF performance criteria, namely \textbf{Proximity}, \textbf{Sparsity} and \textbf{Validity}. Proximity and sparsity assess the similarity or closeness between the original state and its corresponding CF state by quantifying the mean value and number of modified pixels, respectively. Validity is a measure that determines the success rate of the CF generation method by calculating the percentage of generated CF states that successfully alter the model's output to the target action. 
Additionally, we consider \textit{generative} metrics found in the literature that measure the visual authenticity of the generated CF states. These metrics include \textbf{FID} (Fréchet Inception Distance), which assesses the similarity distance between generated and real states based on their feature representations and \textbf{LPIPS} (Learned Perceptual Image Patch Similarity), which evaluates perceptual differences between real and fake state images using high-level visual features.
% , and \textbf{KID} (Kernel Inception Distance), which measures the dissimilarity between feature distributions of generated and real images using the Maximum Mean Discrepancy method. 
% Lastly, by comparing the generated CF states with real-world images, the Inception Score (\textbf{IS}) provides a measure of the quality and diversity exhibited by the counterfactual examples.
Such metrics offer insights into the visual fidelity and realism of the provided CF examples.

\begin{figure*}
\centering
\includegraphics[width=\linewidth, height=5.6cm]{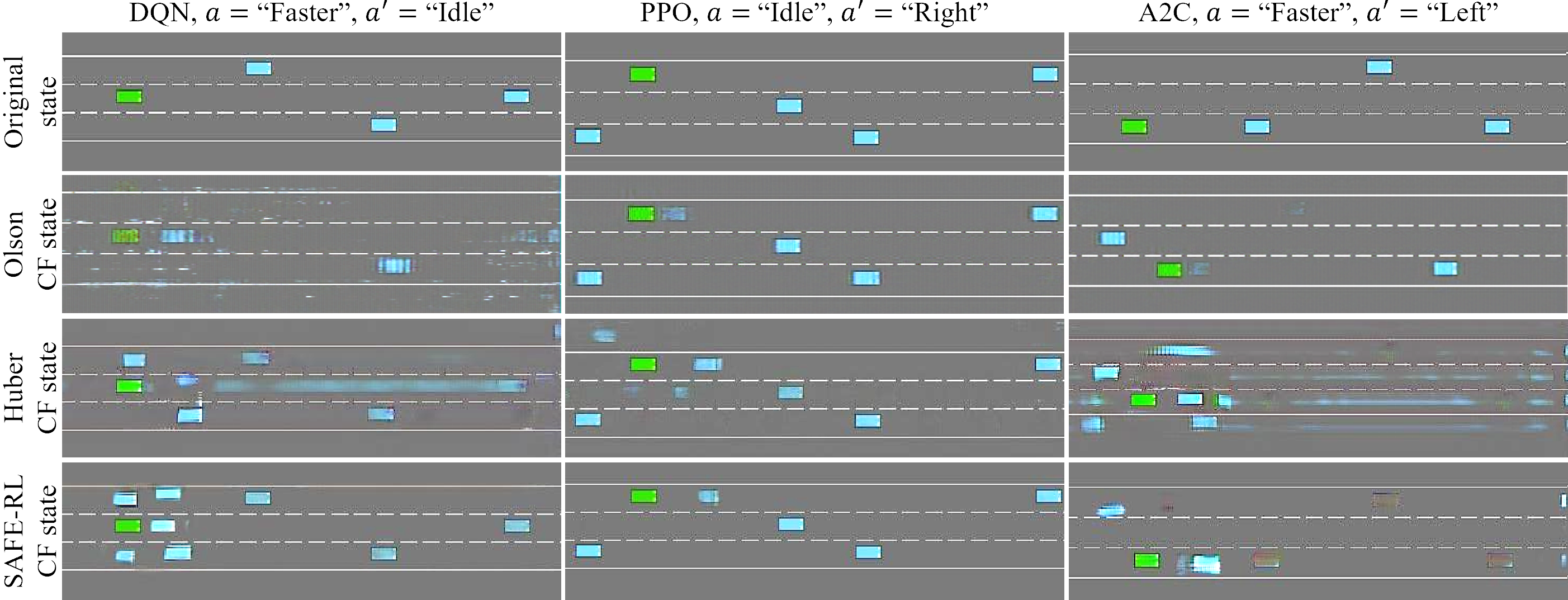}%, height=6cm
\vspace{-0.1cm}
\caption{Visual evaluation of the comparative performance between the SOTA approach and SAFE-RL across three distinct DRL agents: DQN, PPO, and A2C. The initial action is represented as $a$, while the desired action is indicated as $a'$. The EV is portrayed as a green box, situated alongside PVs (blue boxes).}
\label{fig: highway}
% \end{adjustbox}
\vspace{-0.2cm}
\end{figure*}

% \begin{figure}[t!]
% \centering
% \includegraphics[width=\linewidth]{img/highway3.png}
% \caption{Visual evaluation of the comparative performance between the SOTA approach and SAFE-RL across three distinct DRL agents: DQN, PPO, and A2C. The initial action is represented as $a$, while the desired action is indicated as $a'$. The EV is portrayed as a green box, situated alongside PV denoted by blue boxes.}
% \label{fig: highway}
% % \end{adjustbox}
% \vspace{-0.5cm}
% \end{figure}

% \begin{figure}[t!]
% % \vspace{-0.5cm}
% % \begin{adjustbox}{width=\linewidth,center}
% \centering
% \includegraphics[width=\linewidth]{img/roundabout.png}
% \caption{Comparing the visual performance of the SOTA approach and SAFE-RL in the context of a roundabout environment. See the caption of Fig.~\ref{fig: highway} for more details.} %three distinct DRL agents: DQN, PPO, and A2C. The initial action is represented as $a$, while the intended action is denoted by $a'$. The EV is depicted as a green box, positioned alongside PV indicated by blue boxes.}
% \label{fig: roundabout}
% % \end{adjustbox}
% \vspace{-0.4cm}
% \end{figure}

\begin{figure}
\centering
% \vspace{-0.6cm}
\includegraphics[width=\linewidth, height=7cm]{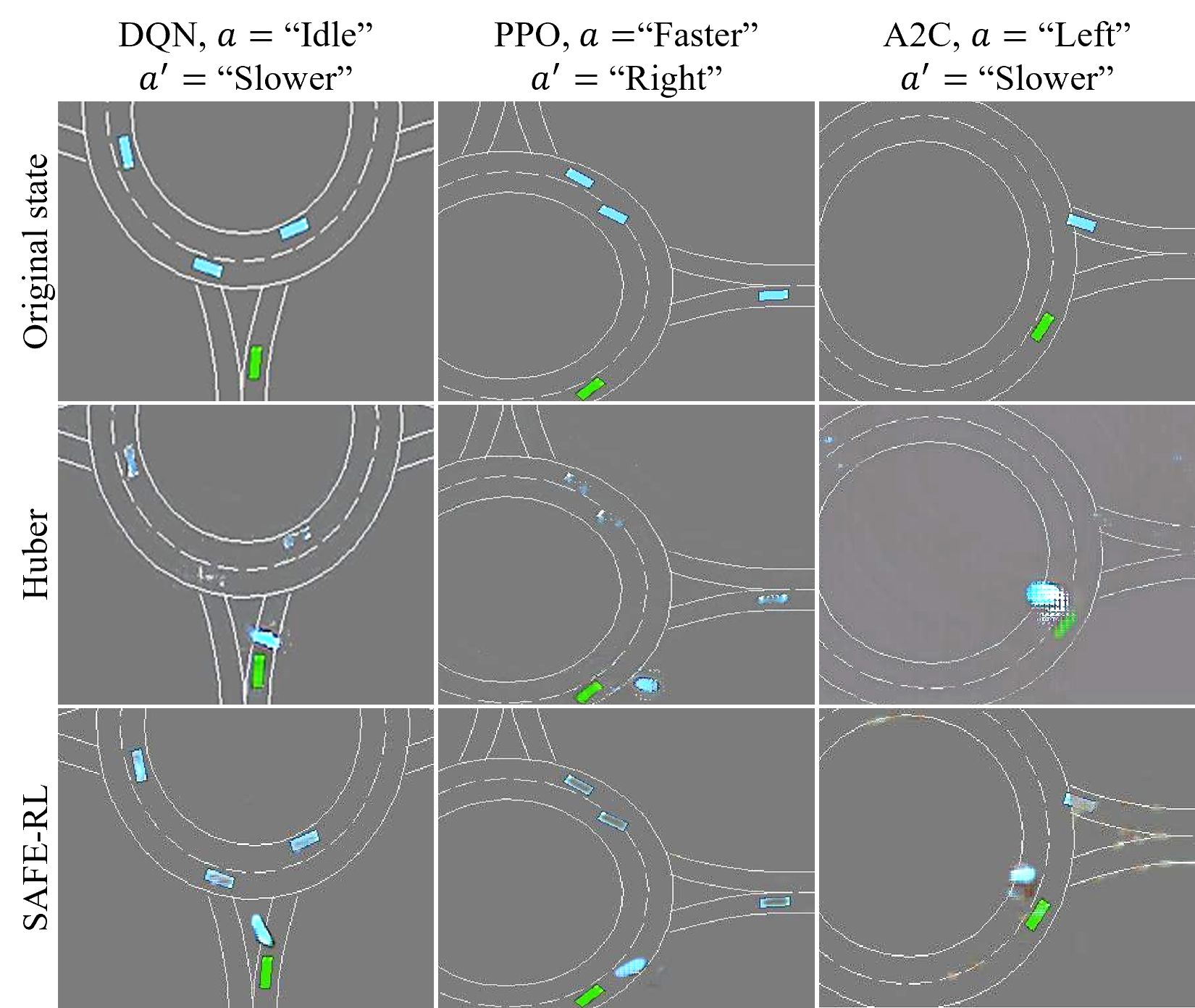} %height=8.5cm
\vspace{-0.5cm}
\caption{A visual comparison of the Huber and SAFE-RL models across three different DRL agents: DQN, PPO, and A2C. The original action is shown by $a$, while the desired actions are marked as $a'$. The EV is depicted as a green box, positioned alongside PVs (blue boxes).}
\label{fig: roudnabout}
% \end{adjustbox}
\vspace{-0.2cm}
\end{figure}

\begin{figure}
\centering
\includegraphics[width=\linewidth, height=9.5cm]{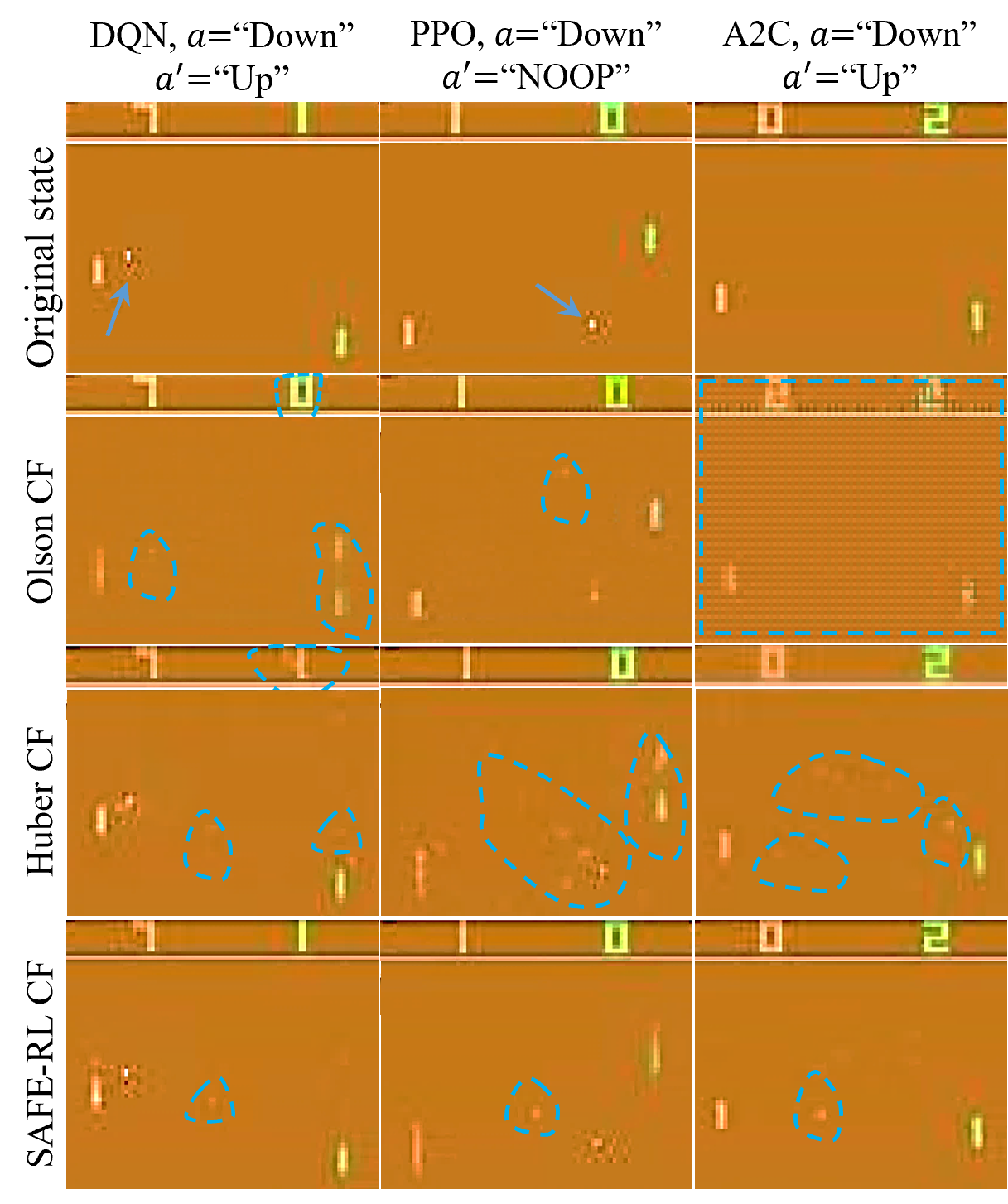}
\vspace{-0.2cm}
\caption{A visual assessment of the comparative performance between the SOTA methods and SAFE-RL across DRL agents: DQN, PPO, and A2C. The initial action is represented as $a$, and the desired action is indicated by $a'$. The blue dashed lines emphasise the modified regions of the query states. For clarity, only the most recent states (input and CFs) are shown, while the trajectory of the ball's movement indicated by blue arrows over the time horizon.}
\label{fig: pong}
\vspace{-0.5cm}
% \end{adjustbox}
\end{figure}

\subsection{Highway, Roundabout and Atari Game\DIFadd{s} Environments}
\label{sec: environments}
The evaluation environments employed in our study encompass the domain of ADS and the Atari Pong game. In the ADS application, we focus on two distinct driving scenarios: Highway (Fig.~\ref{fig: highway}) and Roundabout (Fig.~\ref{fig: roudnabout}). In these scenarios, our pre-trained DRL agents control an EGO Vehicle~(EV) navigating a multi-lane highway or approaching a roundabout amidst flowing traffic. The objective is to drive at the fastest possible pace while avoiding collisions with other Participant Vehicles~(PVs). 
% To encourage the agents to achieve this objective, the following reward function is shaped:
% \begin{equation}\label{eq:reward}
% R(\mathbf{s},a) = \frac{v-v_{min}}{v_{max}-v_{min}} - \beta \times {\text{collision}},
% \end{equation}
% \noindent where $(\mathbf{s}, a)$ denotes the state-action pair,  $\beta$ is a regularisation parameter, $v_{max}$, $v_{min}$, and $v$ are the maximum, minimum, and current velocity of the EV, respectively and the event signal  '${\text{collision}}$' becomes equal to one in the case of collision incidents. 
% %The agents are trained to make informed decisions and map the state $\mathbf{s}$ to action by maximising the cumulative reward. 
The action space consists of five decisions: ``Left'' lane change, ``Right'' lane change, ``Idle'' (lane keeping), ``Faster'' (accelerate), and ``Slower'' (decelerate).

In a different task, the DRL agents are trained to outperform a built-in opponent in the Atari game\DIFadd{s}~\cite{1606.01540}\DIFadd{, which include Pong, Space Invaders and Ms Pacman}. The setup\DIFadd{ of the Pong game}, consists of two paddles: the opponent's paddle displayed in orange on the left, and the agent's paddle displayed in green on the right. The objective of the agent is to redirect the ball from its own side 
% (the right goal) 
and aim it towards the opponent's left goal. The agent has three available actions 
% to manipulate its paddle: 
``NOOP'' (no movement), ``Up'' (move upwards) and ``Down'' (move downwards). With each successful attempt to pass the ball into the opponent's goal, the agent earns a positive point, whereas if the ball goes into its own goal, a negative point is received. \DIFadd{In Space Invaders, the aim is to destroy the invading spaceships using laser cannons controlled by the DRL agent before they reach Earth. The DRL agent earns points for destroying the spaceships, but loses points if any of them make it to Earth. Finally, in Ms. Pacman, the agent should collect all the pellets on the screen while avoiding capture by the ghosts (see~\cite{1606.01540}).}
% for more details).}

\subsection{Performance Evaluation}
\label{sec:PE}
Table~\ref{tab: performance} provides the performance comparison between SAFE-RL and two state-of-art (SOTA) CF explainer models by~\cite{olson2021counterfactual} and~\cite{huber2023ganterfactual} across three environments and three DRL agents. %Upon evaluation of the results from the 27 experiments detailed in Table~\ref{tab: performance}, 
It becomes apparent that SAFE-RL \DIFadd{holds a notable} advantage over the SOTA in terms of the widely acknowledged CF metrics including proximity, sparsity and validity. \DIFadd{Comparing the SAFE-RL with the method proposed by Huber~{\it{et al.}}} \DIFadd{t}he integration of saliency maps in\DIFadd{to our} deep generative model proved \DIFadd{to be} contributory in guiding the generator to introduce targeted modifications to specific pixels\DIFadd{. This resulted} in a reduction in the \DIFadd{both} quantity (sparsity) and the value of altered pixels (proximity) \DIFadd{in most cases}.  The model \DIFadd{proposed by Olson~{\it{et al.}}}~\cite{olson2021counterfactual} managed to achieve \DIFadd{validity} results comparable to SAFE-RL~\DIFadd{in certain instances}\DIFadd{. H}owever, due to the lack of constraints on the solution space, \DIFadd{they have applied more changes, as reflected in higher proximity and sparsity values. Particularly, we observed overfitting of the Olson~{\it{et al.}} model in the SpaceInvaders and MsPacman environments, where it successfully changed the label of the DRL model in all test cases but generated CF states showing repeated scenes. Moreover,} it often \DIFadd{led to the generation of} implausible CF states. This phenomenon will be further discussed in the following section.

Despite the remarkable advancements observed in CF metrics, SAFE-RL demonstrates only marginal improvement in generative metrics. FID and LPIPS, which measure the feature representation and perceptual similarity between input and CF state images, assess the realism of the generated CF images. 
%Notably, the IS metric evaluates the quality and diversity of generated images in relation to the distribution of real-world images. 
% Given that all the input images originate from simulation environments, the relatively low IS values, evaluating the quality of generated CF images in relation to real-world images, are well-founded. Conversely, 
% The FID and LPIPS metrics, measuring the feature representation and perceptual similarity between input and CF state images, underscore that SAFE-RL, with a narrow margin, surpasses the SOTA in performance.

\DIFadd{Table II shows the ablation analysis of SAFE-RL in the Highway Environment for the DQN agent, examining the impact of the introduced loss functions by varying the coefficients $\lambda_{fuse}$ and $\lambda_{pred}$. When both coefficients are zero, SAFE-RL behaves similarly to the Huber \textit{et al.} model. The addition of the ``prediction loss function'' improves validity performance, as seen in the comparison between the first/second and third/fourth rows. However, other metrics remain relatively unchanged. Conversely, utilising the ``fuse loss function'' compels SAFE-RL to focus changes on salient pixels only, resulting in sparser alterations. However, the validity metric does not significantly improve.}

\begin{figure}
% \begin{adjustbox}{width=\linewidth,center}

\centering
\includegraphics[width=\linewidth, height=4.1cm]{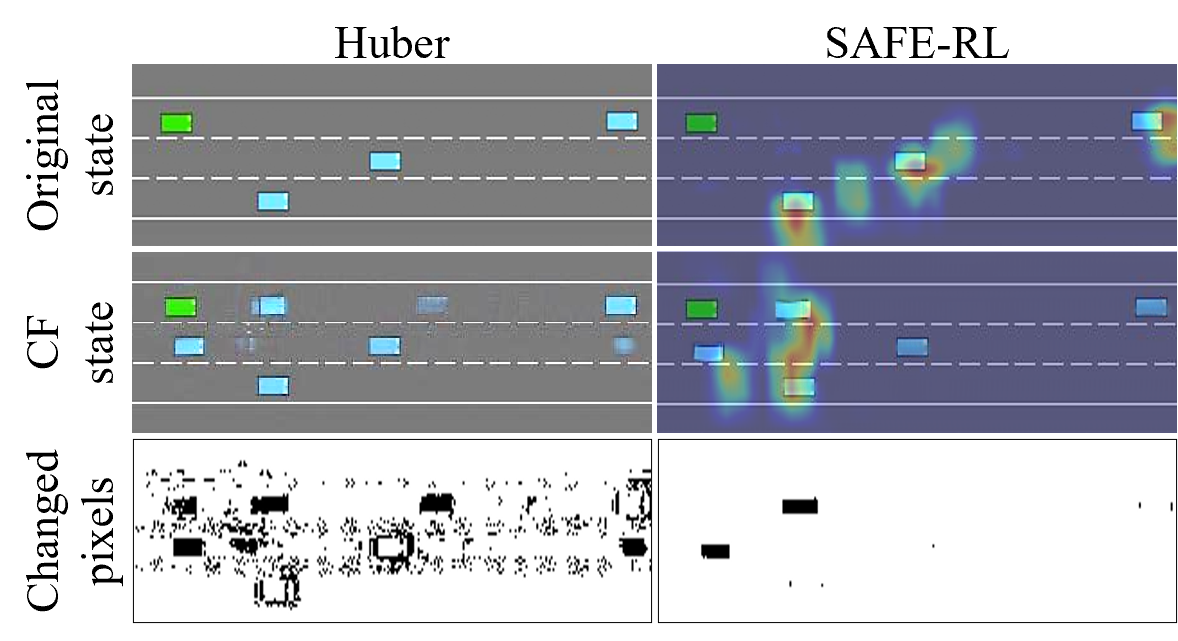}
\vspace{-0.2cm}
\caption{Contrasting the performance of Huber {\it{et al.}} method with the SAFE-RL in highway environment for the DQN agent by illustrating the altered pixels in the third row.} % The top two rows portray the query and CF states, while the third row visualises the disparity between them.}
\label{fig: highway-diff}
% \end{adjustbox}
\vspace{-0.1cm}
\end{figure}

\begin{figure}
% \begin{adjustbox}{width=\linewidth,center}
\centering
\includegraphics[width=\linewidth, height=6.1cm]{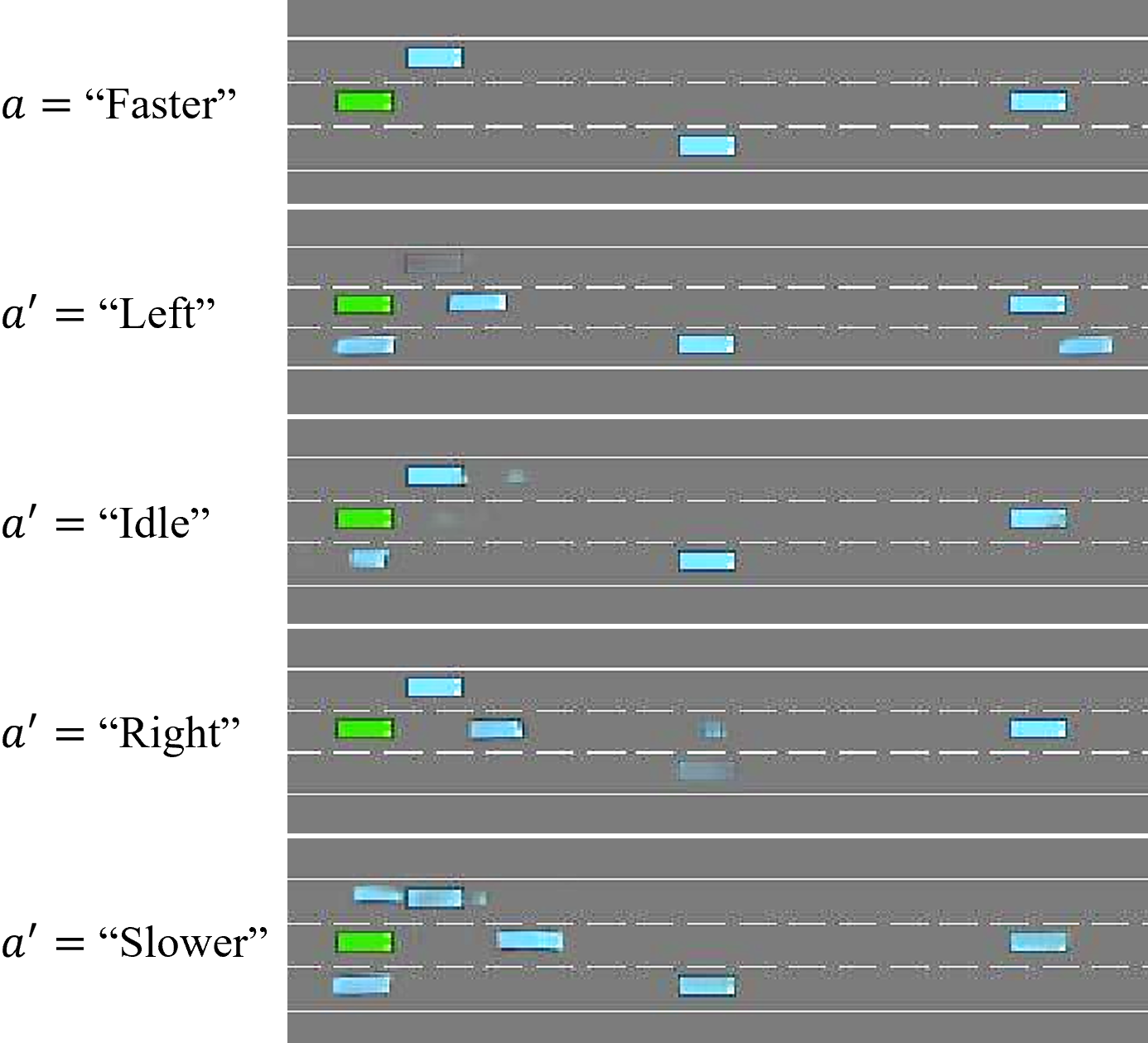}
% \vspace{-0.3cm}
\caption{CF explanations generated by the SAFE-RL model for the DQN agent, covering all potential alternative actions in response to the initial "Faster" decision.} % within a highway driving scenario.}
\label{fig:highway-DQN}
% \end{adjustbox}
\vspace{-0.2cm}
\end{figure}

\subsection{Quality of Counterfactual Explanations}
\label{sec: Quality}

Fig.~\ref{fig: highway} - Fig.~\ref{fig: pong} provide visual comparisons illustrating the perceptible disparity in the quality of generated CF examples between SAFE-RL and the SOTA CF explainer models. Several scenarios are depicted offering insights into the efficacy of the models in generating informative and plausible CF explanations.

To begin with, in the highway environment, for altering the PPO agent's decision from ``Idle'' to  ``Right'' (Fig.~\ref{fig: highway}, second column), the SAFE-RL has added a PV (blue box) ahead of the EV (green box). On the contrary, the Huber {\it{et al.}} method yields an implausible modification by moving a PV off-road to the left. Evidently, that method introduces unrealistic artifacts, clearly visible in all columns of Fig.~\ref{fig: highway}. The model by~\cite{olson2021counterfactual} works better than that of Huber {\it{et al.}}, however, one can see in Fig.~\ref{fig: highway} that the DQN agent does not preclude the possibility of lane-change as the SAFE-RL does.

In the roundabout driving scenario (as depicted in Fig.~\ref{fig: roudnabout}), SAFE-RL's effectiveness is also evident. For instance, when shifting the decision of a DQN agent from "Idle" to "Slower" (as shown in the first column of Fig.~\ref{fig: roudnabout}), SAFE-RL introduces a PV that appears realistic and is positioned ahead of the EV. In contrast, the Huber\textit{et al.} approach fades out irrelevant PVs within the roundabout and introduces implausible modifications, such as moving a PV off-road in the second column and generating an unrealistic PV in the third column. Notably, Olson~\textit{et al.}'s model~\cite{olson2021counterfactual}, due to the high dimensionality of the state images, was unable to converge in the Roundabout environment. However, their results in other environments are reported, depicted and discussed. 

We have witnessed a good quality of CF explanations by the SAFE-RL method in the Atari Pong game environment too; for instance, considering an initial state classified as ``Down'' by the DQN model (Fig.~\ref{fig: pong}, first column), the SAFE-RL model employs a saliency map to produce a modified version wherein a new ball is positioned closer to our paddle as indicated by the dashed blue lines, leading to a decision change to ``Up'' (Fig.~\ref{fig: pong}, fourth row). Conversely, the SOTA models struggle with challenges in achieving such distinct adjustments. Olson {\it{et al.}}'s method yields two paddles, fades the ball and introduces unrelated alterations, such as changing the score points from 7-0 to 7-1 (second row). Huber {\it{et al.}}'s method introduces several irrelevant alterations too (third row). Overall, the SAFE-RL model consistently exhibits sparser changes, further highlighting its superior performance in generating substantively meaningful CF explanations compared to the SOTA methods across all examined scenarios. \DIFadd{Due to space limitations, we have placed the visual results of the Space Invaders and Ms Pacman environments on our ~\href{https://github.com/Amir-Samadi/SAFE-RL}{GitHub} page.}

In a further visual assessment, Fig.~\ref{fig: highway-diff} illustrates how SAFE-RL achieves sparser alterations compared to Huber et al.~\cite{huber2023ganterfactual}. Given the initial query state at the top row, SAFE-RL modifies only the salient regions, as shown in the third row, highlighting the difference between the original and CF states. The non-salient pixels have not been altered in the SAFE-RL model, while on the contrary, the Huber {\it{et al.}} method applied extensive changes along the road. 

\DIFadd{We have already demonstrated quantitatively and qualitatively that focusing on altering the salient pixels has aided SAFE-RL in generating sparser, more realistic, feasible and meaningful CF examples compared to the baseline models. At the same time, comprehensive user studies conducted by Huber~{\it{et al.}}~\cite{huber2023ganterfactual} and Olson~{\it{et al.}}~\cite{olson2021counterfactual} have shown the capability of CF examples in providing understandability for end-users. As a result, SAFE-RL can also offer comprehensive explanations for the end-users. This comes from the fact that CF examples with fewer artifacts (i.e., computationally sparser CF examples) offer better comparisons between the original and CF examples, resulting in better understanding and traceability for end-users~\cite{sharma2022faster}. Moreover, actionable and feasible CF examples could provide more understanding for end-users, as they are more imaginable~\cite{sharma2022faster}. Although conducting a user study falls outside the scope of this research, we} elucidate how CF examples can provide understandable interpretation for end-users and developers.
 
 \DIFadd{Fig.~\ref{fig:highway-DQN}} depict\DIFadd{s} the generated explanations by the SAFE-RL model for the DQN agent in all possible counter-actions.  
 Given the initial state associated with the action ``Faster'' in the top row, the SAFE-RL has applied the following changes for each desired action:
\begin{itemize}
 \item ``Left'': closing the front and right lanes with PV.    
 \item ``Idle'': closing the right lane, as the EV is already at its top speed and the action probability for ``Idle'' and ``Faster'' decisions are equal in the DQN agent.
 \item ``Right'': closing the left and front lanes with PV.     
 \item ``Slower'': surrounding the EV in all lanes.
\end{itemize}

The above comprehensible and informative changes provided by the SAFE-RL, imply that the underlying policy of the DQN agent is rational, as it necessitates reasonable adjustments for decision alterations.

%% file: tex/conclusion.tex
This paper provides quantitative and qualitative evidence that integrating saliency maps within a deep generative adversarial network can enhance the explainability of Deep Reinforcement Learning~(DRL) agents operating in visual environments. The proposed SAFE-RL framework is a stepping stone towards improved transparency and trust in DRL agents that is crucial for deploying them safely in real-world applications like automated driving systems. Enabling human operators to understand the rationale behind the agent's decisions through CF reasoning can bridge the gap between model opacity and human-centric explicability\DIFadd{, as evidenced by the extensive user studies conducted by Olson~{\it{et al.}}~\cite{olson2021counterfactual} and Huber~{\it{et al.}}~\cite{huber2023ganterfactual}. Although a user study was not within the scope of our work, we have discussed how the concepts of sparsity and feasibility, central to our computational approach, contribute to enhancing end-user understandability. By generating sparse and more practical CF examples, SAFE-RL holds promise in providing informative explanations for end-users.} Future work can focus on extending SAFE-RL to handle more complicated observation spaces with higher resolutions across longer time horizons. Overall, SAFE-RL demonstrates that it is possible to generate insightful CF explanations that can elucidate the behaviour of DRL policies to users.

% This paper designed a novel method named after SAFE for generating counterfactual (CF) explanations leveraging a cycle-GAN guided by saliency maps. 
% The latter was adopted to highlight the most salient regions of the input image that play a decisive role in determining the output of the machine learning model and subsequently help alter only those regions given the target (complement) label. The proposed method significantly advanced the state-of-art models in terms of  validity (the higher, the better) and sparsity (the lower, the better) of the generated CF examples using the Berkeley DeepDrive dataset. At the same time, the generated CFs were more interpretable and clear to understand by visual inspection. We believe that this method has the potential to strengthen our trust and facilitate the adoption of machine learning models in real-world applications, such as autonomous vehicles and automated driving systems in general, by providing more transparent and interpretable explanations for their  predictions and decision-making. In the future, it is worth validating the performance of SAFE with other datasets too. 